\documentclass[10pt]{article}
\usepackage{design_ASC}
\usepackage{comment}

\title{\LARGE \bf Online Baum-Welch algorithm for Hierarchical Imitation Learning} 

\author{Vittorio Giammarino\thanks{V.~Giammarino is with Division of Systems Engineering, Boston University, Boston, MA 02446, USA.
{\tt\small vittoriogiammarino@gmail.com}} and Ioannis Ch. Paschalidis\thanks{Ioannis Ch. Paschalidis is with Dept. of Electrical and Computer Engineering, Division of Systems Engineering, and Dept. of Biomedical Engineering, Boston University, 8 St. Mary's St., Boston, MA 02215, USA.
\tt\small yannisp@bu.edu}
}

\date{}
\usepackage{hyperref}
\begin{document}
\setlength{\droptitle}{-5em}    
\maketitle


\begin{abstract}

The options framework for hierarchical reinforcement learning has increased its popularity in recent years and has made improvements in tackling the scalability problem in reinforcement learning. Yet, most of these recent successes are linked with a proper options initialization or discovery. When an expert is available, the options discovery problem can be addressed by learning an options-type hierarchical policy directly from expert demonstrations. This problem is referred to as hierarchical imitation learning and can be handled as an inference problem in a Hidden Markov Model, which is done via an Expectation-Maximization type algorithm. In this work, we propose a novel online algorithm to perform hierarchical imitation learning in the options framework. Further, we discuss the benefits of such an algorithm and compare it with its batch version in classical reinforcement learning benchmarks. We show that this approach works well in both discrete and continuous environments and, under certain conditions, it outperforms the batch version.

\end{abstract}



\section{Introduction}

\begin{figure}
    \centering
    \includegraphics[scale=0.7]{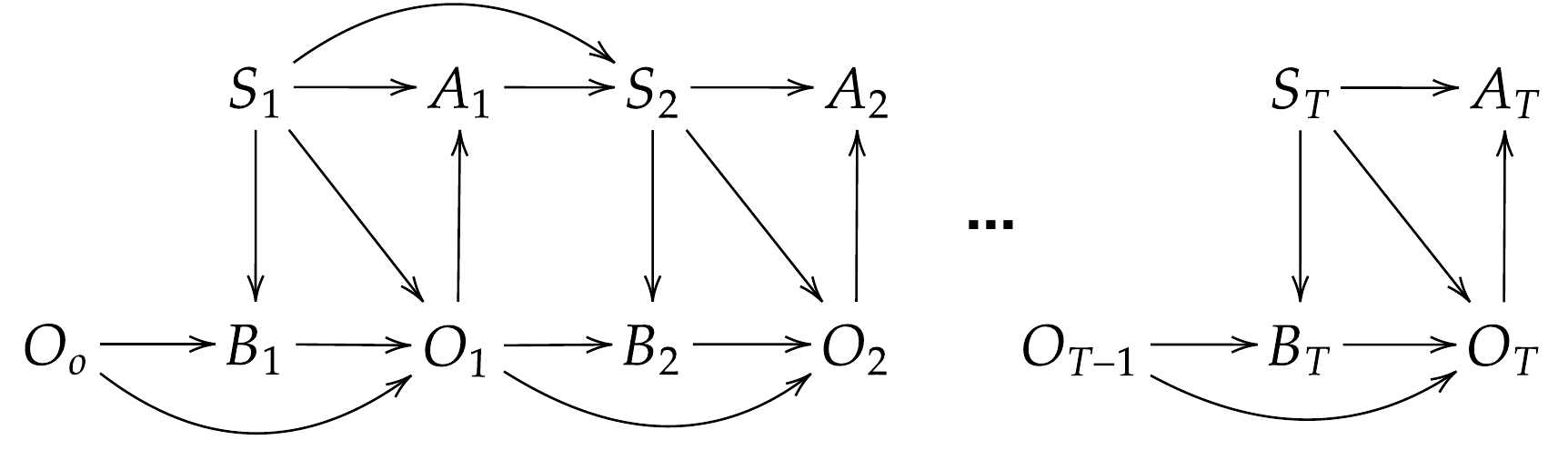}
    \caption{Graphical model for the options framework.}
    \label{fig:option_diag}
\end{figure}

Hierarchical Reinforcement Learning (HRL) addresses the scalability problem in classical Reinforcement Learning (RL) \cite{sutton2018reinforcement} by introducing different levels of temporal abstractions i.e., by dividing the agent policy in decisions that are temporally extended over several steps (higher-level) and in others taken at each step (lower-level). 
Most of the recent successes of HRL (see \cite{nachum2018data} for instance) rely in learning a good hierarchical structure which divides the main problems in sub-problems and tackles them separately by means of single \emph{options} \cite{sutton1999between}. In the literature, the hierarchical learning problem is either decoupled in option initialization, also called option discovery, and in optimal option selection \cite{heess2016learning, kulkarni2016hierarchical, vezhnevets2017feudal, florensa2017stochastic,peng2019mcp}, or it is performed in and end-to-end fashion where the entire hierarchy is learnt while solving the task \cite{bacon2017option, vezhnevets2016strategic}.
When for a specific task an expert is available, initializing policies by direct observation of the expert behavior leads to faster learning convergence \cite{cheng2018fast}. The procedure of learning policies from expert data is called imitation learning \cite{ross2010efficient,ross2011reduction} and, as an extension to HRL, recent works have focused on inferring not only the expert policy but also its underlying hierarchical structure. These studies are generally divided in: Hierarchical Inverse Reinforcement Learning (HIRL), which infers a hierarchical reward function either from expert demonstrations (state-action pairs) \cite{krishnan2016hirl}, or only observations (states) \cite{henderson2017optiongan}; and Hierarchical Imitation Learning, (HIL) which directly learns the expert policy in a hierarchical fashion \cite{le2018hierarchical, yu2018one, fox2019multi, sharma2019third}.
In this paper, we assume that the expert follows an options-type hierarchical policy and we formulate an online algorithm to perform end-to-end HIL. We leverage the idea that the {\em Options Probabilistic Graphical Model (OPGM)} in Fig.~\ref{fig:option_diag} can be handled as a special case of a {\em Hidden Markov Model (HMM)} \cite{rabiner1989tutorial,barto2003recent} and that inference in HMM can be performed via an {\em Expectation-Maximization (EM)} recursion, also known as the {\em Baum-Welch (BW)} algorithm \cite{baum1967inequality, baum1970maximization}. Given the expert demonstrations, this algorithm alternates between an Expectation step (E-step), which computes a surrogate of the log-likelihood, and a Maximization step (M-step), which maximizes such a function over the policy space. By alternating the E-step and the M-step several times, the BW algotrithm is able to find a policy which (locally) maximizes the log-likelihood.\\

\emph{Related Work and Contributions:} Works related to our method are \cite{daniel2016probabilistic, zhang2020provable,fox2017multi}, which exploit a batch version of the BW algorithm to perform end-to-end HIL. In these algorithms, the E-step is carried out through a forward-backward recursion \cite{baum1972inequality}, which needs a sweep through the entire data set at each iteration. As such, for environments where many training samples are required, this procedure is expensive and motivates the development of an online algorithm which processes the data on-the-fly. In addition to efficiency, online algorithms are also memory-wise efficient, since, at each iteration, a single sample is processed and then discarded. Note that we examine the two algorithms in competition; however, batch and online versions are complementary and in practical applications can be used together in sequence. We now summarize the main contributions of this work: $(i)$ the batch version of the BW algorithm for HIL in \cite{daniel2016probabilistic, zhang2020provable, fox2017multi} requires to process the entire data set at each E-step; to tackle this issue, inspired by the works in \cite{mongillo2008online,cappe2011online} for the HMM setting, we develop an online recursion for the OPGM in Fig.~\ref{fig:option_diag} which processes the data on-the-fly. To the best of our knowledge, this is the first online end-to-end algorithm for HIL. $(ii)$ Both \cite{mongillo2008online} and \cite{cappe2011online} make assumptions on the policy parameterization in their recursion for HMM, we try to relax these assumptions for the sake of using non-smooth functions approximations, such as Neural Networks (NN), to parameterize the hierarchical policy. $(iii)$ We compare the two versions of the BW for HIL algorithm via empirical experiments on classical OpenAi RL benchmarks \cite{brockman2016openai}.\footnote{All the code is available at \url{https://github.com/VittorioGiammarino/Online_BWforHIL}.} \\

\emph{Outline:} In Section~\ref{sec:Preliminary}, we introduce the OPGM and the imitation learning problem. Section~\ref{sec:batchBW} introduces the batch BW as in \cite{daniel2016probabilistic, zhang2020provable,fox2017multi} and in Section~\ref{sec:OnlineBW} we formulate the recursion for the online BW for HIL and provide an overview of the algorithm. Finally, Section~\ref{sec:regularizer} presents the regularization penalties we add to the cost function in order to obtain versatile options and Section~\ref{sec:comparison} compares empirically online and batch versions.  \\

\emph{Notation: }We use uppercase letters (e.g., $S_t$) for random variables, lowercase letters (e.g., $s_t$) for values of random variables, script letters (e.g., $\mathcal{S}$) for sets, and bold lowercase letters (e.g., $\bm{\theta}$) for vectors. Let $[t_1 : t_2]$ be the set of integers $t$ such that $t_1 \leq t \leq t_2$; we write $S_t$ such that $t_1 \leq t \leq t_2$ as $S_{t_1 : t_2}$. Moreover, we refer to $\One[S_t = s_t]$ as the indicator function, which is $1$ when $S_t = s_t$ and zero otherwise, and to $\delta(\cdot)$ as the Kronecker delta. Finally, $\mathbb{E}[\cdot]$ represents expectation, $\mathbb{P}(\cdot)$ probability, $|\mathcal{S}|$ the cardinality of a set, and $||\cdot||_2$ the $\ell_2$-norm. 

\section{Preliminary}
\label{sec:Preliminary}
In the following we introduce the OPGM as illustrated in Fig.~\ref{fig:option_diag} and the imitation learning problem.
The index $t$ represents time and $(S_t, A_t, O_t, B_t)$ denote the state, action, option and termination indicator at time $t$, respectively. For all $t$, $S_t$ is defined on the set of states $\mathcal{S}$, possibly infinite, $A_t$ and $O_t$ are respectively defined on the set of actions $\mathcal{A}$ and the set of options $\mathcal{O}$, both finite, and $B_t$ is defined on the binary set $\mathcal{B}=\{0,1\}$. Moreover, define the parameter $\bm{\theta} := (\bm{\theta}_{hi} \in \varTheta_{hi}, \bm{\theta}_{lo} \in \varTheta_{lo}, \bm{\theta}_b \in \varTheta_b)$ where $\varTheta := (\varTheta_{hi} \times \varTheta_{lo} \times \varTheta_b) \subset \mathbb{R}^d$. Given any $(O_0, S_1) = (o_0, s_1)$, the joint distribution on the rest of the OPGM is determined by the following components: an unknown environment transition probability function $P : \mathcal{S} \times \mathcal{A} \rightarrow \Delta_{\mathcal{S}}$ where $\Delta_{\mathcal{S}}$ denotes the space of probability distributions over $\mathcal{S}$, and a triplet of stationary policies $\{\pi_{hi}^{\bm{\theta}_{hi}}, \pi_{lo}^{\bm{\theta}_{lo}}, \pi_{b}^{\bm{\theta}_{b}}\}$ where $\pi_{hi}^{\bm{\theta}_{hi}}$ is the high level policy parameterized by $\bm{\theta}_{hi}$, $\pi_{lo}^{\bm{\theta}_{lo}}$ the low level policy parameterized by $\bm{\theta}_{lo}$ and $\pi_{b}^{\bm{\theta}_{b}}$ the termination policy parameterized by $\bm{\theta}_{b}$. The hierarchical decision process starts at $t=0$, where the agent decides whether to terminate or not the current option $o_0$. This decision is encoded in the termination indicator $b_1$ sampled from $\pi_{b}^{\bm{\theta}_{b}}(\cdot|s_1,o_{0})$, where $\pi_{b}^{\bm{\theta}_{b}}:\mathcal{S}\times \mathcal{O} \rightarrow \Delta_{\mathcal{B}}$. If $b_1 = 1$, the option $o_0$ terminates and the next sample $o_1$ is sampled from $\pi_{hi}^{\bm{\theta}_{hi}}(\cdot|s_1)$, where $\pi_{hi}^{\bm{\theta}_{hi}}: \mathcal{S} \rightarrow \Delta_{\mathcal{O}}$; otherwise, if $b_1=0$, the option $o_0$ continues and $o_1 = o_0$. Next, the action $a_1$ is sampled from $\pi_{lo}^{\bm{\theta}_{lo}}(\cdot|s_1,o_1)$, where $\pi_{lo}^{\bm{\theta}_{lo}}: \mathcal{S} \times \mathcal{O} \rightarrow \Delta_{\mathcal{A}}$, and the agent interacts with the environment through the low level policy associated with the option $o_1$. Finally, the next state $s_2$ is sampled from $P(\cdot|s_1,a_1)$, and the rest of the samples $(s_{3:T}, a_{2:T}, o_{2:T}, b_{2:T})$ are generated analogously. The just described decision process, based on the triplet $\{\pi_{hi}^{\bm{\theta}_{hi}}, \pi_{lo}^{\bm{\theta}_{lo}}, \pi_{b}^{\bm{\theta}_{b}}\}$, encodes the hierarchical agent policy in the options framework. For the sake of completeness, we define $\tilde \pi_{hi}^{\bm{\theta}_{hi}}$ as
\begin{equation}
    \tilde \pi_{hi}^{\bm{\theta}_{hi}}(o_t|o_{t-1},s_t,b_t) := \begin{cases} \pi_{hi}^{\bm{\theta}_{hi}}(o_t|s_t), & \text{if } b_t =1, \\
    1 , & \text{if } b_t=0, o_t=o_{t-1},\\
    0, & \text{if } b_t=0, o_t \neq o_{t-1}.\end{cases}
    \label{eq:tilde_pi}
\end{equation}
Fixing the initial state $S_1 = s_1$ and the initial option $O_0 = o_0$, the joint distribution of $\{S_{2:T}, A_{1:T}, O_{1:T}, B_{1:T}\}$ becomes
\begin{align}
    \begin{split}
        &\mathbb{P}^{\bm{\theta}}_{o_0, s_1}(S_{2:T} = s_{2:T}, A_{1:T} = a_{1:T}, O_{1:T} = o_{1:T}, B_{1:T} = b_{1:T}) = \\
        &\bigg[\prod_{t=1}^T \pi_{b}^{ \bm{\theta}_b}(b_t|s_t,o_{t-1})\tilde\pi_{hi}^{ \bm{\theta}_{hi}}(o_t|o_{t-1},s_t,b_t)\pi_{lo}^{\bm{\theta}_{lo}}(a_t|s_t,o_t) \bigg]\bigg[\prod_{t=1}^{T-1}P(s_{t+1}|s_t,a_t)\bigg].
    \end{split}
    \label{eq:Preliminary_joint_distribution}
\end{align}

Concerning the Imitation Learning (IL) problem, it is defined as inferring the underlying expert distribution via a set of demonstrations (state-action samples) generated while solving a task \cite{hussein2017imitation}. When we assume the expert behavior follows a hierarchical policy with true parameters $(\bm{\theta}^*_{hi}, \bm{\theta}^*_{lo}, \bm{\theta}^*_b)$, and given initial conditions $(o_0,s_1)$, the process of estimating $(\bm{\theta}^*_{hi}, \bm{\theta}^*_{lo}, \bm{\theta}^*_b)$ through a finite sequence of expert demonstrations $(s_{2:T},a_{1:T})$ with $T \geq 2$ is called HIL. One way to address this problem is by solving:
\begin{equation}
    \max_{(\bm{\theta}_{hi}, \bm{\theta}_{lo}, \bm{\theta}_b) \in \varTheta}  \mathcal{L}(\bm{\theta}_{hi}, \bm{\theta}_{lo}, \bm{\theta}_b),
    \label{eq:IL_max_prob}
\end{equation}
where $\mathcal{L}(\bm{\theta}_{hi}, \bm{\theta}_{lo}, \bm{\theta}_b)$ denotes the marginal log-likelihood and is equivalent to the logarithm of the joint probability of generating the expert demonstrations $(s_{2:T},a_{1:T})$ given $(o_0,s_1)$ and the parameters $\bm{\theta}_{hi}, \bm{\theta}_{lo}, \bm{\theta}_b$, i.e.,
\begin{equation}
    \mathcal{L}(\bm{\theta}_{hi}, \bm{\theta}_{lo}, \bm{\theta}_b) = \log \mathbb{P}^{\bm{\theta}}_{o_0,s_1}(s_{2:T},a_{1:T}) = \log \sum_{o_{1:T},b_{1:T}} \mathbb{P}^{\bm{\theta}}_{o_0,s_1}(s_{2:T}, a_{1:T}, o_{1:T}, b_{1:T}). 
    \label{eq:log_likelihood}
\end{equation}
Note that, $\mathbb{P}^{\bm{\theta}}_{o_0,s_1}(s_{2:T}, a_{1:T}, o_{1:T}, b_{1:T})$ in \eqref{eq:log_likelihood} is the same as \eqref{eq:Preliminary_joint_distribution}, but we have dropped the random variables $S_{2:T}, A_{1:T}, O_{1:T}, B_{1:T}$ to streamline the notation.
The optimization problem in \eqref{eq:IL_max_prob} is hard to evaluate for our framework, considering that for a long sequence of demonstrations the cost function in \eqref{eq:log_likelihood} gets close to zero. Yet, the BW algorithm provides an iterative procedure based on EM which solves \eqref{eq:IL_max_prob} by maximizing a surrogate of \eqref{eq:log_likelihood}. The way we compute this surrogate during the E-step determines the main difference between the batch and our online version of the algorithm.    

\section{Batch Baum-Welch for Hierarchical Imitation Learning}
\label{sec:batchBW}
In this section we draw the main ingredients of the batch BW for HIL as in \cite{zhang2020provable}. As mentioned, this algorithm alternates between the E-step and the M-step: during the E-step we compute a surrogate of \eqref{eq:log_likelihood}, the Baum's auxiliary function \cite{baum1970maximization}, with respect to the previously obtained vector of parameters $\bm{\theta^{\text{old}}}$. Then, in the M-step, we optimize this function with respect to a new vector of parameters $\bm{\theta}$. Given $(O_0,S_1) = (o_0, s_1)$, we obtain the following Baum's auxiliary function for the OPGM (cf. Appendix~\ref{App_HMM:derivation_of_Q} for the complete derivation)
\begin{equation}
    Q_{o_0,s_1}^{T}(\bm{\theta} | \bm{\theta}^{\text{old}}) = \frac{1}{T}    \sum_{o_{1:T},b_{1:T}}\mathbb{P}^{\bm{\theta^{\text{old}}}}_{o_0,s_1}(o_{1:T}, b_{1:T}|(s_t,a_t)_{1:T})\log \mathbb{P}^{\bm{\theta}}_{o_0,s_1}(s_{2:T},a_{1:T}, o_{1:T}, b_{1:T}).
    \label{eq:Q_BW_opt}
\end{equation}
By replacing $\mathbb{P}^{\bm{\theta}}_{o_0,s_1}(s_{2:T},a_{1:T}, o_{1:T}, b_{1:T})$ with \eqref{eq:Preliminary_joint_distribution}, Eq.~\eqref{eq:Q_BW_opt} becomes
\begin{align}
    \begin{split}
    &Q_{o_0,s_1}^{T}(\bm{\theta} | \bm{\theta}^{\text{old}}) = \frac{1}{T}\bigg\{\sum_{t=2}^T\sum_{o_{t-1}}\sum_{b_t}\mathbb{P}^{\bm{\theta}^{\text{old}}}_{o_0,s_1}(o_{t-1},b_t|(s_t,a_t)_{1:T})\log \pi_{b}^{\bm{\theta}_b}(b_t|s_t, o_{t-1}) \\
    &+ \sum_{t=1}^T\sum_{o_{t-1}}\sum_{b_t}\sum_{o_t}\mathbb{P}^{\bm{\theta}^{\text{old}}}_{o_0,s_1}(o_{t-1},b_t,o_t|(s_t,a_t)_{1:T})\log \tilde \pi_{hi}^{\bm{\theta}_{hi}}(o_t|o_{t-1},s_t,b_t) \\
    &+\sum_{t=1}^T\sum_{o_t}\sum_{b_t}\mathbb{P}^{\bm{\theta}^{\text{old}}}_{o_0,s_1}(o_t,b_t|(s_t,a_t)_{1:T})\log \pi_{lo}^{\bm{\theta}_{lo}}(a_t|s_t,o_t) 
    + \sum_{b_1}\mathbb{P}^{\bm{\theta}^{\text{old}}}_{o_0,s_1}(b_1|(s_t,a_t)_{1:T})\log \pi_{b}^{\bm{\theta}_b}(b_1|s_1, o_{0}) + C \bigg\},
    \label{eq:Q_BW_opt_interm}
    \end{split}
\end{align}
where $C$ is constant with respect to $\bm{\theta}$.
In \eqref{eq:Q_BW_opt_interm}, $\mathbb{P}^{\bm{\theta}^{\text{old}}}_{o_0,s_1}(o_{t-1},b_t|(s_t,a_t)_{1:T})$ and $\mathbb{P}^{\bm{\theta}^{\text{old}}}_{o_0,s_1}(o_t,b_t|(s_t,a_t)_{1:T})$ are referred to as the smoothing distributions of the latent variables given the expert demonstrations $(s_t,a_t)_{1:T}$ and are computed, during the E-step of the batch algorithm, via forward-backward decomposition (cf.~Appendix~\ref{app:Forward-Backward} and \cite{zhang2020provable}). Moreover, $C$ contains all constant terms (independent on $\bm{\theta}$), \\ $\mathbb{P}^{\bm{\theta}^{\text{old}}}_{o_0,s_1}(b_1|(s_t,a_t)_{1:T})\log \pi_{b}^{\bm{\theta}_b}(b_1|s_1, o_{0})$ is neglected, for $T$ large enough, for reasons linked with the forward-backward decomposition \cite{zhang2020provable}, and $\tilde \pi_{hi}^{\bm{\theta}_{hi}}$ depends on ${\bm{\theta}_{hi}}$ only through $\pi_{hi}^{\bm{\theta}_{hi}}$ in \eqref{eq:tilde_pi} for $b_t=1$. Hence, using the convention $0\log0 = 0$, we can replace $Q_{o_0,s_1}^{T}(\bm{\theta} | \bm{\theta}^{\text{old}})$ by
\begin{align}
\begin{split}
    &Q_{o_0,s_1}^{T}(\bm{\theta} | \bm{\theta}^{\text{old}}) = \frac{1}{T}\bigg\{\sum_{t=2}^T\sum_{o_{t-1}}\sum_{b_t}\mathbb{P}^{\bm{\theta}^{\text{old}}}_{o_0,s_1}(o_{t-1},b_t|(s_t,a_t)_{1:T})\log \pi_{b}^{\bm{\theta}_b}(b_t|s_t, o_{t-1}) \\
    &+ \sum_{t=1}^T\sum_{o_t}\mathbb{P}^{\bm{\theta}^{\text{old}}}_{o_0,s_1}(o_t,b_t=1|(s_t,a_t)_{1:T})\log \pi_{hi}^{\bm{\theta}_{hi}}(o_t|s_t)
    +\sum_{t=1}^T\sum_{o_t}\sum_{b_t}\mathbb{P}^{\bm{\theta}^{\text{old}}}_{o_0,s_1}(o_t,b_t|(s_t,a_t)_{1:T})\log \pi_{lo}^{\bm{\theta}_{lo}}(a_t|s_t,o_t)\bigg\}.
    \label{eq:Q_BW_opt_final}
\end{split}
\end{align}

We summarize the batch BW for HIL recursion in Algorithm~\ref{alg:batchBW}. As discussed, the main shortcoming of this algorithm is the need of processing the entire set of demonstrations, multiple times, at each iteration.   

\begin{algorithm}
\caption{Batch Baum-Welch algorithm for HIL}
\begin{algorithmic}[1]
\State \textbf{Require:} Observation sequence $(s_t,a_t)_{1:T}$; $o_0 \in \mathcal{O}$; $s_1 \in \mathcal{S}$; $N \in \mathbb{N}_{+}$ and $\bm{\theta}^{(0)} \in \varTheta$.
\For{$n=1, \dots, N$}
\State Compute $\{\mathbb{P}^{\bm{\theta}^{(n-1)}}_{o_0,s_1}(o_{t-1},b_t|(s_t,a_t)_{1:T})\}_{t=2}^T$ and $\{\mathbb{P}^{\bm{\theta}^{(n-1)}}_{o_0,s_1}(o_t,b_t|(s_t,a_t)_{1:T})\}_{t=1}^T$ \Comment{E-step}
\State Update $\bm{\theta}^{(n)} \in \arg\max_{\bm{\theta} \in \Theta}Q_{o_0,s_1}^{T}(\bm{\theta} |\bm{\theta}^{(n-1)})$ based on \eqref{eq:Q_BW_opt_final} \Comment{M-step}
\EndFor
\end{algorithmic}
\label{alg:batchBW}
\end{algorithm}

\section{Online Baum-Welch for Hierarchical Imitation Learning}
\label{sec:OnlineBW}
In the following, we replace the smoothing distributions $\mathbb{P}^{\bm{\theta}^{\text{old}}}_{o_0,s_1}(o_{t-1},b_t|(s_t,a_t)_{1:T})$ and $\mathbb{P}^{\bm{\theta}^{\text{old}}}_{o_0,s_1}(o_t,b_t|(s_t,a_t)_{1:T})$ in \eqref{eq:Q_BW_opt_final} with a sufficient statistic $\phi_{T}^{\bm{\theta}}$ which is updated as soon as the new state-action pair $(s_{T},a_{T})$ becomes available. For simplicity, we make the following assumption.
\begin{assumption}\label{Ass_2}
State $S_t$ and action $A_t$ take their values in a finite set, $\mathcal{S}$ and $\mathcal{A}$ respectively.
\end{assumption}
Given $O_0=o_0$ and $S_1=s_1$, the sufficient statistic $\phi_{T}^{\bm{\theta}}:\mathcal{O}^2\times \mathcal{B}\times\mathcal{\tilde A}\times\mathcal{\tilde S} \rightarrow \mathbb{R}$, where $\mathcal{O}$ is the set of options, $\mathcal{B}$ is the termination binary set, and $\mathcal{\tilde A} \subseteq \mathcal{A}$ and $\mathcal{\tilde S} \subseteq \mathcal{S}$ are respectively the set of actions and states explored by the expert, is defined as:
\begin{align}
    \begin{split}
        \phi_{T}^{\bm{\theta}}(o', b, o, s, a) = 
        \frac{1}{T}\mathbb{E}_{o_0,s_1}^{\bm{\theta}}\bigg[ \sum_{t=1}^T\One[O_{t-1} = o', B_t = b, O_t = o, S_t=s, A_t=a] \bigg| (s_t,a_t)_{1:T}  \bigg].
    \end{split}
    \label{eq:phi}
\end{align}
Note that, to avoid confusion and distinguish between $o_t$ as the value of the random variable $O_t$ at time $t$ and $o_t$ as an element of the set $\mathcal{O}$, we change the notation compared to Section~\ref{sec:batchBW}. Therefore, in \eqref{eq:phi} we use $o',o \in \mathcal{O}$, $b \in \mathcal{B}$, $s \in \mathcal{\tilde S}$, $a \in \mathcal{\tilde A}$ while we keep the notation $(s_t,a_t)_{1:T}$ for the expert demonstrations. In Proposition~\ref{prop:M-step_ass2}, a new Baum's auxiliary function for the online BW for HIL is obtained in terms of \eqref{eq:phi}.
\begin{prop}
\label{prop:M-step_ass2}
Under Assumption~\ref{Ass_2} we can write \eqref{eq:Q_BW_opt} as
\begin{equation}
    \begin{split}
    Q_{o_0,s_1}^{T}(\bm{\theta} | \bm{\theta}^{\text{old}}) =& \sum_{o'}\sum_{o}\sum_s\sum_a \Big\{\sum_{b}\phi_{T}^{\bm{\theta}^{\text{old}}}(o', b, o, s, a)\log\pi_{b}^{\bm{\theta}_b}(b|s, o') \\
    &+\phi_{T}^{\bm{\theta}^{\text{old}}}(o', b = 1, o, s, a)\log \pi_{hi}^{\bm{\theta}_{hi}}(o|s) + \sum_{b}\phi_{T}^{\bm{\theta}^{\text{old}}}(o', b, o, s, a)\log \pi_{lo}^{\bm{\theta}_{lo}}(a|s,o)\Big\},
    \label{eq:Q_online_BW_Final}
    \end{split}
\end{equation}

where $\pi_{hi}^{\bm{\theta}_{hi}}(o|s)$, $\pi_{lo}^{\bm{\theta}_{lo}}(a|s,o)$ and $\pi_{b}^{\bm{\theta}_b}(b|s, o)$ are parameterized by tabular parameterization and $\phi_{T}^{\bm{\theta}}$ is defined in \eqref{eq:phi}. 
\begin{proof}
For the long version see Appendix~\ref{app:OnlineBW_Mstep}. We recall the Baum's auxiliary function for the OPGM in \eqref{eq:Q_BW_opt_interm} where we change the notation as explained earlier. By using the total probability law with respect to $O_{t-1}=o'$ and $O_t=o$ (when necessary) and neglecting $C$ we rewrite Eq.~\eqref{eq:Q_BW_opt_interm} more compactly as 
\begin{align*}
    \begin{split}
    &Q_{o_0,s_1}^{T}(\bm{\theta} | \bm{\theta}^{\text{old}}) = \frac{1}{T}\sum_{t=1}^T\sum_{o'}\sum_{b}\sum_{o}\mathbb{P}^{\bm{\theta}^{\text{old}}}_{o_0,s_1}(O_{t-1} = o', B_t = b, O_t = o|(s_t,a_t)_{1:T})\\
    &\times \log\pi_{b}^{\bm{\theta}_b}(B_t = b|S_t = s_t, O_{t-1}=o')\tilde \pi_{hi}^{\bm{\theta}_{hi}}(O_t = o|O_{t-1}=o',S_t = s_t,B_t = b)\pi_{lo}^{\bm{\theta}_{lo}}(A_t=a_t|S_t=s_t,O_t = o).
    \end{split}
\end{align*}
Note that, we write $S_t=s_t$ and $A_t=a_t$ to emphasize that in the previous equation $(s_t,a_t)$ is still the expert demonstration at time $t$. Then, we proceed as follows  
\begin{align*}
     &Q_{o_0,s_1}^{T}(\bm{\theta} | \bm{\theta}^{\text{old}}) =\sum_{o'}\sum_b\sum_o\frac{1}{T}\sum_{t=1}^T\mathbb{P}^{\bm{\theta}^{\text{old}}}_{o_0,s_1}(O_{t-1}=o',B_t=b,O_t=o|(s_t,a_t)_{1:T})\sum_s\sum_a\delta(s-s_t)\delta(a-a_t) \\ &\times \log\pi_{b}^{\bm{\theta}_b}(B_t = b|S_t = s, O_{t-1}=o')\tilde \pi_{hi}^{\bm{\theta}_{hi}}(O_t = o|O_{t-1}=o',S_t = s, B_t = b)\pi_{lo}^{\bm{\theta}_{lo}}(A_t=a|S_t=s,O_t = o)\\
     =&\sum_{o'}\sum_b\sum_o\sum_s\sum_a\frac{1}{T}\sum_{t=1}^T\mathbb{P}^{\bm{\theta}^{\text{old}}}_{o_0,s_1}(O_{t-1}=o',B_t=b,O_t=o,S_t=s,A_t=a|(s_t,a_t)_{1:T}) \\
     &\times \log\pi_{b}^{\bm{\theta}_b}(B_t = b|S_t = s, O_{t-1}=o')\tilde \pi_{hi}^{\bm{\theta}_{hi}}(O_t = o|O_{t-1}=o',S_t = s, B_t = b)\pi_{lo}^{\bm{\theta}_{lo}}(A_t=a|S_t=s,O_t = o) \\
     =&\sum_{o'}\sum_b\sum_o\sum_s\sum_a\frac{1}{T}\mathbb{E}_{o_0,s_1}^{\bm{\theta}^{\text{old}}}\bigg[ \sum_{t=1}^T\One[O_{t-1} = o', B_t = b, O_t = o, S_t=s, A_t=a] \bigg| (s_t,a_t)_{1:T}\bigg]\\
     &\times \log\pi_{b}^{\bm{\theta}_b}(B_t = b|S_t = s, O_{t-1}=o')\tilde \pi_{hi}^{\bm{\theta}_{hi}}(O_t = o|O_{t-1}=o',S_t = s, B_t = b)\pi_{lo}^{\bm{\theta}_{lo}}(A_t=a|S_t=s,O_t = o),
\end{align*}
and \eqref{eq:Q_online_BW_Final} follows.
\end{proof}
\end{prop}

\begin{remark}
In practice, Assumption~\ref{Ass_2} is not crucial and can be relaxed to allow for $\pi_{hi}^{\bm{\theta}_{hi}}$, $\pi_{lo}^{\bm{\theta}_{lo}}$, $\pi_{b}^{\bm{\theta}_b}$ and $\phi_{T}^{\bm{\theta}}$ function approximations with neural networks. The new $Q_{o_0,s_1}^{T}(\bm{\theta} | \bm{\theta}^{\text{old}})$ in \eqref{eq:Q_online_BW_Final} needs to sum over $s \in \mathcal{\tilde S}$, $a \in \mathcal{\tilde A}$ in place of the sum over $T$ in \eqref{eq:Q_BW_opt_final}. Note that in many practical situations $|\mathcal{\tilde A}|\times|\mathcal{\tilde S}| < T$.
\end{remark}

We proceed introducing the online recursion to update $\phi_{T}^{\bm{\theta}}$ in \eqref{eq:phi}. First, $\phi_{T}^{\bm{\theta}}$ can be decomposed through the following two filters
\begin{align}
    &\chi_{T}^{\bm{\theta}}(o) = \mathbb{P}^{\bm{\theta}}_{o_0,s_1}\big(O_T=o\big|(s_t,a_t)_{1:T}\big), \label{eq:OnlineBW_chi} \\
    \begin{split}
        &\rho_{T}^{\bm{\theta}}(o',b,o,s,a,o'')
        = \frac{1}{T}\mathbb{E}^{\bm{\theta}}_{o_0,s_1}\bigg[ \sum_{t=1}^T\One[O_{t-1} = o', B_t = b, O_t = o, S_t=s, A_t=a]\bigg| O_T=o'', (s_t, a_t)_{1:T}  \bigg], \label{eq:OnlineBW_rho}
    \end{split}
\end{align}
where $\chi_{T}^{\bm{\theta}}:\mathcal{O} \rightarrow \Delta_{\mathcal{O}}$ and $\rho_{T}^{\bm{\theta}}: \mathcal{O}^3\times\mathcal{B}\times\mathcal{\tilde A}\times\mathcal{\tilde S} \rightarrow \mathbb{R}$.
It follows that
\begin{align}
\begin{split}
    \phi_{T}^{\bm{\theta}}(o',b,o,s,a) &= \sum_{o''}\rho_{T}^{\bm{\theta}}(o',b,o,s,a,o'')\chi_{T}^{\bm{\theta}}(o'').
    \label{eq:phi_onlineBW}
\end{split}
\end{align}
Proposition~\ref{prop:OnlineBW_decomposition} shows the recursion to update \eqref{eq:OnlineBW_chi}-\eqref{eq:OnlineBW_rho}.
\begin{prop}
\label{prop:OnlineBW_decomposition}
\begin{itemize}
    \item \textbf{Initialization:}
\begin{align}
    &\chi_{0}^{\bm{\theta}}(o) = P(O_0 = o), \label{eq:propOBW_chi_init} \\
    &\rho_{0}^{\bm{\theta}}(o', b, o, s, a, o'') = 0,  \label{eq:propOBW_rho_init}
\end{align}
\item \textbf{Recursion:} for $T>0$ and the new state-action pair $(s_T,a_T)$ it holds that
\begin{align}
\begin{split}
        &\chi_{T}^{\bm{\theta}}(o)
        = \sum_{o'}\sum_{b}\frac{\pi_{lo}^{\bm{\theta}_{lo}}(a_T | s_T, o)\tilde \pi_{hi}^{\bm{\theta}_{hi}}(o|b,s_T,o')\pi_b^{\bm{\theta}_{b}}(b|s_T,o')\chi_{T-1}^{\bm{\theta}}(o')}{\sum_{o'}\sum_{b}\sum_{o}\pi_{lo}^{\bm{\theta}_{lo}}(a_T | s_T, o)\tilde \pi_{hi}^{\bm{\theta}_{hi}}(o|b,s_T,o')\pi_b^{\bm{\theta}_{b}}(b|s_T,o')\chi_{T-1}^{\bm{\theta}}(o')},\\
        \label{eq:propOBW_chi}
\end{split}\\
\begin{split}
    &\rho_{T}^{\bm{\theta}}(o', b, o, s, a, o'')
    = \sum_{o'''}\sum_{b'}\bigg\{\frac{1}{T}\kappa(o'-o''', b-b', o-o'', s-s_T, a-a_T) \\ &+  \bigg(1-\frac{1}{T}\bigg)\rho_{T-1}^{\bm{\theta}}(o', b, o, s, a, o''')\bigg\} \frac{\tilde\pi_{hi}^{\bm{\theta}_{hi}}(o''|b',s_T,o''')\pi_b^{\bm{\theta}_{b}}(b'|s_T,o''')\chi_{T-1}^{\bm{\theta}}(o''')}{\sum_{o'''}\sum_{b'}\tilde \pi_{hi}^{\bm{\theta}_{hi}}(o''|b',s_T,o''')\pi_b^{\bm{\theta}_{b}}(b'|s_T,o''')\chi_{T-1}^{\bm{\theta}}(o''')}.
    \label{eq:propOBW_rho}
\end{split}
\end{align}
\end{itemize}
where $\kappa(o'-o''', b-b', o-o'', s-s_T, a-a_T)=\delta(o'-o''')\delta(b-b')\delta(o-o'')\delta(s-s_T)\delta(a-a_T)$.
\begin{proof}
See Appendix~\ref{app:OnlineBW}.
\end{proof}
\end{prop}
Based on Propositions~\ref{prop:M-step_ass2}-\ref{prop:OnlineBW_decomposition}, we formulate Algorithm~\ref{alg:onlineBW_ass2} which is, to the best of our knowledge, the first online EM type algorithm suitable for end-to-end HIL within the options framework.
In Algorithm~\ref{alg:onlineBW_ass2}, note that, we do not have to specify the number of iterations ($N$, in Algorithm~\ref{alg:batchBW}) as we perform an E-step after each state-action pair available. Additionally, we inhibit the M-step for $t < T_{\text{min}}$ to ensure that $Q_{o_0,s_1}^{T}(\bm{\theta} | \bm{\theta}^{\text{old}})$ is numerically well-behaved which is not always the case for a small number of demonstrations.

\begin{algorithm}
\caption{Online Baum-Welch algorithm for HIL}
\begin{algorithmic}[1]
\State \textbf{Require:} Observation sequence $(s_t,a_t)_{1:T}$; $O_0=o$; $S_1=s$; $\theta^{(0)} \in \varTheta$.
\For{$t= 0, \dots, T$}
\If{$t=0$}
\State Initialize $\rho_{0}^{\bm{\theta}^{(0)}}$ and $\chi_{0}^{\bm{\theta}^{(0)}}$ in \eqref{eq:propOBW_rho_init} and \eqref{eq:propOBW_chi_init} respectively \Comment{Initialization}
\EndIf
\If{$t>0$}
\State Compute $\rho_{t}^{\bm{\theta}^{(t-1)}}$ and $\chi_{t}^{\bm{\theta}^{(t-1)}}$ in \eqref{eq:propOBW_rho} and \eqref{eq:propOBW_chi} respectively \Comment{E-step}
\If{$t>T_{\text{min}}$}
    \State Compute $\phi_{t}^{\bm{\theta}^{(t-1)}}$ in \eqref{eq:phi_onlineBW}
    \State Update $\bm{\theta}^{(t)} \in \arg\max_{\bm{\theta}^{(t)} \in \varTheta}Q_{o_0,s_1}^{T}(\bm{\theta}^{(t)} | \bm{\theta}^{(t-1)})$ with $Q_{o_0,s_1}^{T}(\bm{\theta} | \bm{\theta}^{\text{old}})$ in \eqref{eq:Q_online_BW_Final} \Comment{M-step}
    \Else 
    \State $\bm{\theta}^{(t)}=\bm{\theta}^{(t-1)}$
\EndIf
\EndIf
\EndFor
\end{algorithmic}
\label{alg:onlineBW_ass2}
\end{algorithm}

\section{Regularization Penalties}
\label{sec:regularizer}
As additional requirement in HIL, we want to learn a set of interpretable and transferable options. To achieve this goal, we penalize the Baum's auxiliary functions in \eqref{eq:Q_BW_opt_final} and \eqref{eq:Q_online_BW_Final} with regularizers on both the high and low level policies \cite{henderson2017optiongan}.

\emph{High level policy regularizers:} For $\pi_{hi}^{\bm{\theta}_{hi}}$, we introduce two regularizers $L_b$ and $L_v$ in \eqref{eq:pi_hi_reg}. By minimizing $L_b$, we encourage the activation of each option with a target sparsity value $\tau = 1/|\mathcal{O}|$ in expectation over the training set. On the other hand, maximizing $L_v$, where var denotes the variance, we encourage the options activation to be varied and force each option to have a high probability for certain states and low for the rest.
\begin{align}
    L_b = \sum_{o} \big|\big| \mathbb{E}_{s}[\pi_{hi}^{\bm{\theta}_{hi}}(o|s)] - \tau \big|\big|_{2}, \ \ \ \ \ \
    L_v = \sum_{o} \text{var}_{s}[\pi_{hi}^{\bm{\theta}_{hi}}(o|s)].  \label{eq:pi_hi_reg}
\end{align}

\emph{Low level policy regularizer:} Additionally, we maximize the Kullback–Leibler divergence ($D_{KL}$) of each low level policy over the set of demonstrations \eqref{eq:regularizer_KL_div}. This in order to enhance differentiation in $\pi_{lo}^{\bm{\theta}_{lo}}$ given different options:
    \begin{align}
    \begin{split}
      L_{D_{KL}} &= \sum_{o}\sum_{o',o\neq o'} D_{KL}\big(\pi_{lo}^{\bm{\theta}_{lo}}(a|s,o)||\pi_{lo}^{\bm{\theta}_{lo}}(a|s,o')\big).
    \end{split}
    \label{eq:regularizer_KL_div}
    \end{align}
Overall, at each M-step, we solve the following optimization problem
\begin{equation}
    \bm{\theta}^{(T)} \in \arg\max_{\bm{\theta}^{(T)} \in \varTheta}Q_{o_0,s_1}^{T}(\bm{\theta} | \bm{\theta}^{\text{old}}) - \lambda_b L_b + \lambda_v L_v + \lambda_{D_{KL}}L_{D_{KL}}.
\end{equation}
Note that, given the different ways we construct the Baum's auxiliary functions for Algorithm~\ref{alg:batchBW} and \ref{alg:onlineBW_ass2}, respectively in \eqref{eq:Q_BW_opt_final} and \eqref{eq:Q_online_BW_Final}, the three penalties are differently implemented in the two settings. For more details on this regard refer to the Appendix~\ref{app:regularizers}.

\section{Comparison and Discussion}
\label{sec:comparison}

\subsection{Implementation and Numerical Complexity}
\label{sec:complexity}
In the following we focus on the numerical complexity of the E-step for both Algorithms~\ref{alg:batchBW} and \ref{alg:onlineBW_ass2}, since this is where the two algorithms differ the most. Algorithm~\ref{alg:batchBW} uses the forward-backward decomposition (Appendix~\ref{app:Forward-Backward}): at each iteration updates a vector of size $T \times |\mathcal{O}| \times |\mathcal{B}|$ in $O(T \times |\mathcal{O}|^2 \times |\mathcal{B}|)$. On the other hand, Algorithm~\ref{alg:onlineBW_ass2} updates a vector of dimension $|\mathcal{\tilde S}| \times |\mathcal{\tilde A}| \times |\mathcal{O}|^2 \times |\mathcal{B}|$ with a numerical complexity of $O(|\mathcal{\tilde S}| \times |\mathcal{\tilde A}| \times |\mathcal{O}|^3 \times |\mathcal{B}|^2)$. The bottleneck of the batch algorithm is $T$, the size of the expert demonstrations; while, in the online algorithm it is $|\mathcal{\tilde S}| \times |\mathcal{\tilde A}|$, which is the combination of states explored and actions used by the expert. Generally, in order to obtain satisfactory learning in stochastic environments we have $T>>|\mathcal{\tilde S}| \times |\mathcal{\tilde A}|$ which implies that a single online E-step in Proposition~\ref{prop:OnlineBW_decomposition} is more efficient than a single forward-backward decomposition. However, consider that the batch BW (Algorithm~\ref{alg:batchBW}) requires $N$ iterations while, the online (Algorithm~\ref{alg:onlineBW_ass2}) $T$ iterations where usually $T>>N$. Therefore, as acknowledged in \cite{cappe2011online,cappe2006inference}, this mere comparison is not always meaningful and requires to be further investigated via empirical experiments.

\subsection{Experiments}
\label{sec:experiments}
\begin{figure}[h!]
    \centering
    \begin{subfigure}[t]{0.45\textwidth}
        \centering
        \includegraphics[width=7cm]{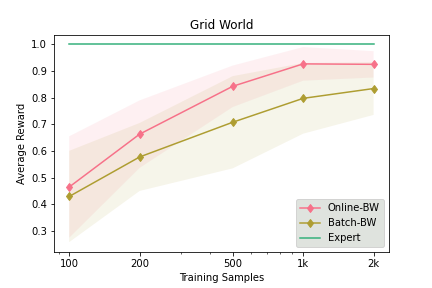}
        \label{fig:Reward_GridWorld_NN}
    \end{subfigure}%
    ~
    \begin{subfigure}[t]{0.45\textwidth}
        \centering
        \includegraphics[width=7cm]{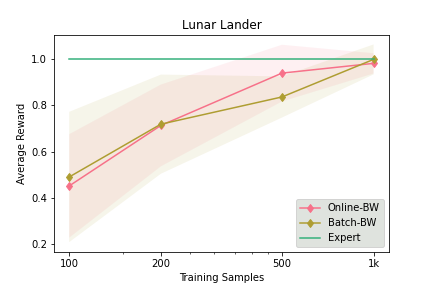}
        \label{fig:Reward_LL}
    \end{subfigure}
    ~
    \begin{subfigure}[t]{0.45\textwidth}
        \centering
        \includegraphics[width=7cm]{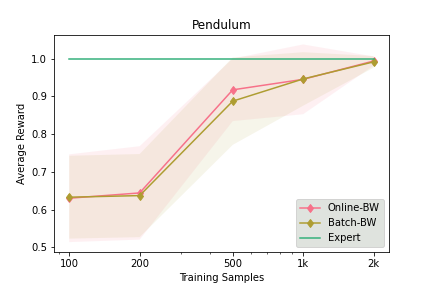}
        \label{fig:Reward_LL}
    \end{subfigure}
    ~
    \begin{subfigure}[t]{0.45\textwidth}
        \centering
        \includegraphics[width=7cm]{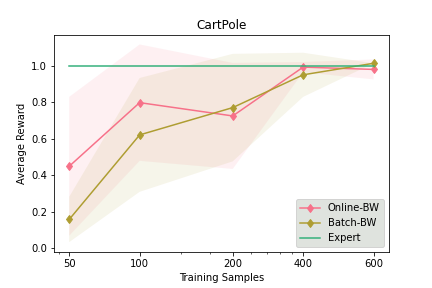}
        \label{fig:Reward_CP}
    \end{subfigure}
    \caption{Performance for different environments. The $y$-axis shows the reward averaged over the random seeds and scaled such that the expert achieves $1$. The $x$-axis shows the size of the training set used for each trial. The shaded area indicates the standard deviation over the 10 reruns for cartpole, lunar lander and pendulum and over 30 reruns for the grid world.}
    \label{fig:Experiments}
\end{figure}

We evaluate the two algorithms on 4 different tasks from the classic RL literature: three of them come from the OpenAI gym library \cite{brockman2016openai}, the cartpole, pendulum and lunar lander; and have a continuous state-space. The last is a grid-world type environment with discrete state-space and high stochasticity in transition and reward. We first generate the expert demonstrations running value iteration on the grid-world, Q-learning on the pendulum and cartpole \cite{sutton2018reinforcement} and a heuristic on lunar lander. After generating the demonstrations, we use Algorithms~\ref{alg:batchBW} and \ref{alg:onlineBW_ass2} to train an options-type hierarchical policy on the same triplet of feed-forward fully connected neural networks. For both $\pi_{lo}^{\bm{\theta}_{lo}}$ and $\pi_{b}^{\bm{\theta}_{b}}$ we use a number of $|\mathcal{O}|=2$ networks made of a single hidden layer of $30$ units, with ReLu activation function; while, $\pi_{hi}^{\bm{\theta}_{hi}}$ uses the same architecture but with $100$ units. All the networks weights are initialized randomly with a uniform distribution $\mathcal{U}(-0.5,0.5)$ at the beginning of each trial. In different trials, the algorithms are fed with a different number of demonstrations (training samples) and they are trained for the same amount of time, on the same hardware and in exactly the same conditions. We run all trials $30$ times over different random seeds for the grid world and $10$ times for the others. After the training is completed, we measure the average reward obtained over $100$ episodes for each trial given a seed and finally, average again over the seeds. More information on the used hyperparameters are provided in Appendix~\ref{app:experiments}. The obtained results are illustrated in Figure~\ref{fig:Experiments}. 

As Figure~\ref{fig:Experiments} depicts, there is no tangible deterioration in the performance when using the online setting with respect to the batch. For the environments with a continuous state space, i.e., lunar lander, pendulum and cart pole, where $T \approx |\mathcal{\tilde S}| \times |\mathcal{\tilde A}|$ we observe similar performance; while, for the grid-world, which has a discrete state space, the online algorithm outperforms its batch version since in this case $T > |\mathcal{\tilde S}| \times |\mathcal{\tilde A}|$.  Finally, note that these experiments are conducted on a reasonably small number of demonstrations $T$, in order to facilitate the comparison, and the training hyperparameters (Appendix \ref{app:regularizers}) are selected such that the two algorithms perform an equivalent amount of gradient steps. For greater $T$, the gap would have been larger since the forward-backward decomposition in the batch algorithm is more expensive. Overall, the results are encouraging and experiments on more realistic setups will be the subject of subsequent work.

\section{Conclusions}
In this work, we develop an online version of the BW algorithm for HIL. Specifically, we formulate an online smoothing recursion suitable for the options framework and leverage it to obtain our online BW algorithm for HIL. In addition, we empirically compare online and batch versions on classical control tasks: the two algorithms show similar performance in all the environments with a continuous state-space; while, when the size of the training set becomes larger compared to the portion of the state-space explored by the expert, e.g the grid-world, we show the online algorithm to be convenient since the forward-backward decomposition used in the batch algorithm becomes more expensive than the online recursion in Proposition~\ref{prop:OnlineBW_decomposition}.

\section{Acknowledgements}
We thank Zhiyu Zhang for all the comments and useful discussions. This work was supported in part by NSF under grants DMS-1664644, CNS-1645681, IIS-1914792, by ARPA-E under grant DE-AR0001282, by the ONR under grant N00014-19-1-2571, and by the NIH under grant R01 GM135930.

\newpage

\printbibliography

\begin{appendix}
\section{Appendix}
\subsection{Derivation of the Baum's auxiliary function}
\label{App_HMM:derivation_of_Q}
We start from the marginal log-likelihood in Eq.~\eqref{eq:log_likelihood}
\begin{equation}
    \mathcal{L}(\bm{\theta}) = \log \mathbb{P}^{\bm{\theta}}_{o_0,s_1}(s_{2:T},a_{1:T}) = \log \sum_{o_{1:T},b_{1:T}} \mathbb{P}^{\bm{\theta}}_{o_0,s_1}(s_{2:T},a_{1:T}, o_{1:T}, b_{1:T}). 
    \label{eq:log_likelihood_appendix}
\end{equation}
For any distribution $q(o_{1:T}, b_{1:T})$ over the hidden states and exploiting the concavity of the logarithm we can obtain a lower bound on \eqref{eq:log_likelihood_appendix} by means of Jensen's inequality:
\begin{align*}
    \begin{split}
    \mathcal{L}(\bm{\theta}) &= \log \sum_{o_{1:T},b_{1:T}} q(o_{1:T}, b_{1:T}) \frac{\mathbb{P}^{\bm{\theta}}_{o_0,s_1}(s_{2:T},a_{1:T}, o_{1:T}, b_{1:T})}{q(o_{1:T}, b_{1:T})}\\
    &\geq \sum_{o_{1:T},b_{1:T}}  q(o_{1:T}, b_{1:T}) \log \frac{\mathbb{P}^{\bm{\theta}}_{o_0,s_1}(s_{2:T},a_{1:T}, o_{1:T}, b_{1:T})}{q(o_{1:T}, b_{1:T})} \equiv F(q,\bm{\theta}).
    \end{split}
\end{align*}
For $F(q,\bm{\theta})$ we then obtain 
\begin{align*}
    F(q,\bm{\theta}) &= \sum_{o_{1:T},b_{1:T}}  q(o_{1:T}, b_{1:T}) \log \frac{\mathbb{P}^{\bm{\theta}}_{o_0,s_1}(s_{2:T}, a_{1:T}, o_{1:T}, b_{1:T})}{q(o_{1:T}, b_{1:T})} \\
    &= \sum_{o_{1:T},b_{1:T}}q(o_{1:T}, b_{1:T})\log\frac{\mathbb{P}^{\bm{\theta}}_{o_0,s_1}(o_{1:T}, b_{1:T}|(s_t,a_t)_{1:T})\mathbb{P}^{\bm{\theta}}_{o_0,s_1}((s_t,a_t)_{1:T})}{q(o_{1:T}, b_{1:T})} \\
    &= \sum_{o_{1:T},b_{1:T}}q(o_{1:T}, b_{1:T}) \log \mathbb{P}^{\bm{\theta}}_{o_0,s_1}((s_t,a_t)_{1:T}) + \sum_{o_{1:T},b_{1:T}}q(o_{1:T}, b_{1:T}) \log \frac{\mathbb{P}^{\bm{\theta}}_{o_0,s_1}(o_{1:T}, b_{1:T}|(s_t,a_t)_{1:T})}{q(o_{1:T}, b_{1:T})}\\
    &= \mathcal{L}(\bm{\theta}) - D_{\text{KL}}[q(o_{1:T}, b_{1:T})||\mathbb{P}^{\bm{\theta}}_{o_0,s_1}(o_{1:T}, b_{1:T}|(s_t,a_t)_{1:T})]. \numberthis \label{eq:HMM_Auxiliary_function_F}
\end{align*}
The second term in \eqref{eq:HMM_Auxiliary_function_F} is the Kullback-Leiber divergence ($D_{KL}$) between the distributions $q(o_{1:T}, b_{1:T})$ and $\mathbb{P}^{\bm{\theta}}_{o_0,s_1}(o_{1:T}, b_{1:T}|(s_t,a_t)_{1:T})$. Therefore, for fixed $\bm{\tilde\theta}$, $F(q,\bm{\theta})$ is maximized when\\
$D_{\text{KL}}[q(o_{1:T}, b_{1:T})||\mathbb{P}^{\bm{\tilde \theta}}_{o_0,s_1}(o_{1:T}, b_{1:T}|(s_t,a_t)_{1:T})]=0$ i.e., $q(o_{1:T}, b_{1:T})=\mathbb{P}^{\bm{\tilde \theta}}_{o_0,s_1}(o_{1:T}, b_{1:T}|(s_t,a_t)_{1:T})$. This yields
\begin{align*}
        F(q,\bm{\theta}) =& \sum_{o_{1:T},b_{1:T}}\mathbb{P}^{\bm{\tilde \theta}}_{o_0,s_1}(o_{1:T}, b_{1:T}|(s_t,a_t)_{1:T}) \log \frac{\mathbb{P}^{\bm{\theta}}_{o_0,s_1}(s_{2:T},a_{1:T}, o_{1:T}, b_{1:T})}{\mathbb{P}^{\bm{\tilde \theta}}_{o_0,s_1}(o_{1:T}, b_{1:T}|(s_t,a_t)_{1:T})} \\
        =& \sum_{o_{1:T},b_{1:T}}\mathbb{P}^{\bm{\tilde \theta}}_{o_0,s_1}(o_{1:T}, b_{1:T}|(s_t,a_t)_{1:T})\log \mathbb{P}^{\bm{\theta}}_{o_0,s_1}(s_{2:T},a_{1:T}, o_{1:T}, b_{1:T}) \\ 
        &- \sum_{o_{1:T},b_{1:T}}\mathbb{P}^{\bm{\tilde \theta}}_{o_0,s_1}(o_{1:T}, b_{1:T}|(s_t,a_t)_{1:T})\log \mathbb{P}^{\bm{\tilde \theta}}_{o_0,s_1}(o_{1:T}, b_{1:T}|(s_t,a_t)_{1:T})).\numberthis \label{eq:F_part_negligible}
\end{align*}
The second term in the last equation in \eqref{eq:F_part_negligible} does not depend on $\bm{\theta}$ and it can be neglected in the optimization problem. Eventually, we end up with the following surrogate of the marginal log-likelihood 
\begin{equation}
    \sum_{o_{1:T},b_{1:T}}\mathbb{P}^{\bm{\tilde \theta}}_{o_0,s_1}(o_{1:T}, b_{1:T}|(s_t,a_t)_{1:T})\log \mathbb{P}^{\bm{\theta}}_{o_0,s_1}(s_{2:T},a_{1:T}, o_{1:T}, b_{1:T}),
    \label{eq:appendix_surrogate}
\end{equation}
and by normalizing \eqref{eq:appendix_surrogate} for numerical stability, we obtain \eqref{eq:Q_BW_opt}.

\subsection{Smoothing via Forward-Backward decomposition for options}
\label{app:Forward-Backward}

In the following, we introduce the Forward-Backward decomposition used by the batch BW to estimate the smoothing distributions. Note that, all these quantities are probability mass functions and require a proper normalizing factor; we use the symbol $\propto$ to denote proportionality. Given $O_0, S_1 = o_0, s_1$, The forward variable $\alpha_t^{\bm{\theta}}(o_t,b_t)$ is defined
\begin{equation}
    \alpha_t^{\bm{\theta}}(o_t,b_t) = \mathbb{P}^{\bm{\theta}}_{o_0,s_1}(A_{1:t} = a_{1:t}, O_t = o_t, B_t = b_t | S_{2:t} = s_{2:t}),
    \label{eq:alpha}
\end{equation}
where in $\alpha_t^{\bm{\theta}}(o_t,b_t)$ we omit the dependency on $S_{2:t} = s_{2:t}, A_{1:t} = a_{1:t}$ and on $o_0, s_1$.
The recursion for \eqref{eq:alpha} becomes
\begin{align*}
    \alpha_t^{\bm{\theta}}(o_t,b_t) \propto 
    \begin{cases}
        \pi_{b}^{ \bm{\theta}_b}(b_t|s_t,o_{t-1})\tilde\pi_{hi}^{ \bm{\theta}_{hi}}(o_t|o_{t-1},s_t,b_t)\pi_{lo}^{\bm{\theta}_{lo}}(a_t|s_t,o_t),  & \text{if } t=1,\\
        \sum_{o_{t-1},b_{t-1}} \pi_{b}^{ \bm{\theta}_b}(b_t|s_t,o_{t-1})\tilde\pi_{hi}^{ \bm{\theta}_{hi}}(o_t|o_{t-1},s_t,b_t)\pi_{lo}^{\bm{\theta}_{lo}}(a_t|s_t,o_t) \alpha_{t-1}^{\bm{\theta}}(o_{t-1},b_{t-1}), & \text{if } 1<t<T.
    \end{cases}
    \label{eq:batchBW_alpha}
\end{align*}
Similarly, the backward variable $\beta_t^{\bm{\theta}}(o_t,b_t)$ is defined
\begin{equation*}
    \beta_t^{\bm{\theta}}(o_t,b_t) =  \mathbb{P}^{\bm{\theta}}_{o_0,s_1}(A_{t+1:T} = a_{t+1:T} | S_{t+1 : T} = s_{t+1 : T}, O_t = o_t, B_t = b_t),
    \label{eq:beta}
\end{equation*}
and its recursion
\begin{align*}
    \beta_t^{\bm{\theta}}(o_t,b_t) \propto \begin{cases}\sum_{o_{t+1},b_{t+1}} \pi_{b}^{\bm{\theta}_b}(b_{t+1}|s_{t+1}, o_t)\tilde\pi_{hi}^{\bm{\theta}_{hi}}(o_{t+1}|o_t,s_{t+1},b_{t+1})\pi_{lo}^{\bm{\theta}_{lo}}(a_{t+1}|s_{t+1},o_{t+1})\beta_{t+1}^{\bm{\theta}}(o_{t+1},b_{t+1}), & \text{if } 1<t<T, \\
    1, & \text{if } t=T.
    \end{cases}
\end{align*}
Then, exploiting forward and backward variables we compute for $\forall t \in [1:T] $ the smoothing distribution $\gamma_t^{\bm{\theta}}(o_t,b_t)$
\begin{align}
\begin{split}
    \gamma_t^{\bm{\theta}}(o_t,b_t) = \mathbb{P}^{\bm{\theta}}_{o_0,s_1}( O_t = o_t, B_t = b_t| S_{2:t} = s_{2:t}, A_{1:t} = a_{1:t}) = \frac{\alpha_t^{\bm{\theta}}(o_t,b_t)\beta_t^{\bm{\theta}}(o_t,b_t)}{\sum_{o_t}\sum_{b_t}\alpha_t^{\bm{\theta}}(o_t,b_t)\beta_t^{\bm{\theta}}(o_t,b_t)},
\end{split}
\label{eq:batchBW_gamma}
\end{align}
and the bi-variate smoothing $\xi_t^{\bm{\theta}}(o_{t-1},b_t)$
\begin{align}
\begin{split}
    &\xi_t^{\bm{\theta}}(o_{t-1},b_t) = \mathbb{P}^{\bm{\theta}}_{o_0,s_1}( O_{t-1} = o_{t-1}, B_t = b_t| S_{2:t} = s_{2:t}, A_{1:t} = a_{1:t})\\
    &\propto \sum_{b_{t-1}}\alpha_{t-1}^{\bm{\theta}}(o_{t-1},b_{t-1})\bigg(\pi_{b}^{\bm{\theta}_b}(b_{t}|s_{t}, o_{t-1})\sum_{o_t}\tilde\pi_{hi}^{\bm{\theta}_{hi}}(o_t|o_{t-1},s_t,b_t)\pi_{lo}^{\bm{\theta}_{lo}}(a_t|s_t,o_t)\beta_t^{\bm{\theta}}(o_t,b_t)\bigg).
\end{split}
\label{eq:batchBW_xi}
\end{align}
Given \eqref{eq:batchBW_gamma}, \eqref{eq:batchBW_xi} and $\tilde\pi_{hi}^{\bm{\theta}_{hi}}$ in \eqref{eq:tilde_pi}, the Baum's auxiliary function in \eqref{eq:Q_BW_opt_final} becomes
\begin{align*}
    \begin{split}
    &Q_{o_0,s_1}^T(\bm{\theta} | \bm{\theta}^{\text{old}}) = \frac{1}{T}\bigg\{\sum_{t=2}^T\sum_{o_{t-1}}\sum_{b_t}\xi_t^{\bm{\theta}^{\text{old}}}(o_{t-1},b_t)\log \pi_{b}^{\bm{\theta}_b}(b_t|s_t, o_{t-1}) + \sum_{t=1}^T\sum_{o_t}\gamma_t^{\bm{\theta}^{\text{old}}}(o_t,b_t=1)\log \pi_{hi}^{\bm{\theta}_{hi}}(o_t|s_t) \\
    &+\sum_{t=1}^T\sum_{o_t}\sum_{b_t}\gamma_t^{\bm{\theta}^{\text{old}}}(o_t,b_t)\log \pi_{lo}^{\bm{\theta}_{lo}}(a_t|s_t,o_t)\bigg\}.
    \end{split}
\end{align*}

\subsection{Proof of Proposition~\ref{prop:M-step_ass2}}
\label{app:OnlineBW_Mstep}
We recall the Baum's auxiliary function for the OPGM in \eqref{eq:Q_BW_opt_interm}
\begin{align*}
    \begin{split}
    &Q_{o_0,s_1}^{T}(\bm{\theta} | \bm{\theta}^{\text{old}}) = \frac{1}{T}\bigg\{\sum_{t=2}^T\sum_{o'}\sum_{b}\mathbb{P}^{\bm{\theta}^{\text{old}}}_{o_0,s_1}(O_{t-1}=o', B_t=b|(s_t,a_t)_{1:T})\log \pi_{b}^{\bm{\theta}_b}(B_t=b|S_t=s_t, O_{t-1}=o') \\
    &+ \sum_{t=1}^T\sum_{o'}\sum_{b}\sum_o\mathbb{P}^{\bm{\theta}^{\text{old}}}_{o_0,s_1}(O_{t-1}=o',B_t=b,O_t=o|(s_t,a_t)_{1:T})\log \tilde \pi_{hi}^{\bm{\theta}_{hi}}(O_t = o|O_{t-1}=o',S_t = s_t,B_t = b) \\
    &+\sum_{t=1}^T\sum_{o}\sum_{b}\mathbb{P}^{\bm{\theta}^{\text{old}}}_{o_0,s_1}(O_t=o,B_t=b|(s_t,a_t)_{1:T})\log \pi_{lo}^{\bm{\theta}_{lo}}(A_t=a_t|S_t=s_t,O_t=o) \\
    &+ \sum_{b}\mathbb{P}^{\bm{\theta}^{\text{old}}}_{o_0,s_1}(B_1=b|(s_t,a_t)_{1:T})\log \pi_{b}^{\bm{\theta}_b}(B_1=b|S_1=s_1, O_0=o_{0}) + C \bigg\}.
    \end{split}
\end{align*}
By using the total probability law with respect to $O_{t-1}=o'$ and $O_t=o$ (when necessary) and neglecting $C$ we rewrite  
\begin{align*}
    \begin{split}
    &Q_{o_0,s_1}^{T}(\bm{\theta} | \bm{\theta}^{\text{old}}) = \frac{1}{T}\bigg\{\sum_{t=2}^T\sum_{o'}\sum_{b}\sum_{o}\mathbb{P}^{\bm{\theta}^{\text{old}}}_{o_0,s_1}(O_{t-1}=o', B_t=b, O_t=o|(s_t,a_t)_{1:T})\log \pi_{b}^{\bm{\theta}_b}(B_t=b|S_t=s_t, O_{t-1}=o') \\
    &+ \sum_{t=1}^T\sum_{o'}\sum_{b}\sum_{o}\mathbb{P}^{\bm{\theta}^{\text{old}}}_{o_0,s_1}(O_{t-1}=o', B_t=b, O_t=o|(s_t,a_t)_{1:T})\log \tilde \pi_{hi}^{\bm{\theta}_{hi}}(O_t = o|O_{t-1}=o',S_t = s_t,B_t = b) \\
    &+\sum_{t=1}^T\sum_{o'}\sum_{b}\sum_{o}\mathbb{P}^{\bm{\theta}^{\text{old}}}_{o_0,s_1}(O_{t-1}=o', B_t=b, O_t=o|(s_t,a_t)_{1:T})\log \pi_{lo}^{\bm{\theta}_{lo}}(A_t=a_t|S_t=s_t,O_t=o) \\
    &+ \sum_{o'}\sum_{b}\sum_{o}\mathbb{P}^{\bm{\theta}^{\text{old}}}_{o_0,s_1}(O_0=o_0,B_1=b,O_1=o_1|(s_t,a_t)_{1:T})\log \pi_{b}^{\bm{\theta}_b}(B_1=b|S_1=s_1, O_0=o_{0})\bigg\}.
    \end{split}
\end{align*}
And then more compactly
\begin{align*}
    \begin{split}
    &Q_{o_0,s_1}^{T}(\bm{\theta} | \bm{\theta}^{\text{old}}) = \frac{1}{T}\sum_{t=1}^T\sum_{o'}\sum_{b}\sum_{o}\mathbb{P}^{\bm{\theta}^{\text{old}}}_{o_0,s_1}(O_{t-1} = o', B_t = b, O_t = o|(s_t,a_t)_{1:T})\\
    &\times \log\pi_{b}^{\bm{\theta}_b}(B_t = b|S_t = s_t, O_{t-1}=o')\tilde \pi_{hi}^{\bm{\theta}_{hi}}(O_t = o|O_{t-1}=o',S_t = s_t,B_t = b)\pi_{lo}^{\bm{\theta}_{lo}}(A_t=a_t|S_t=s_t,O_t = o).
    \end{split}
\end{align*}
Note that, we write $S_t=s_t$ and $A_t=a_t$ to emphasize that in the previous equation $(s_t,a_t)$ is still the expert demonstration at time $t$. We use the sifting property of the Kronecker delta and we obtain  
\begin{align*}
     &Q_{o_0,s_1}^{T}(\bm{\theta} | \bm{\theta}^{\text{old}}) = \sum_{o'}\sum_b\sum_o\frac{1}{T}\sum_{t=1}^T\mathbb{P}^{\bm{\theta}^{\text{old}}}_{o_0,s_1}(O_{t-1}=o',B_t=b,O_t=o|(s_t,a_t)_{1:T})\sum_s\sum_a\delta(s-s_t)\delta(a-a_t), \\
          &\times \log\pi_{b}^{\bm{\theta}_b}(B_t = b|S_t = s, O_{t-1}=o')\tilde \pi_{hi}^{\bm{\theta}_{hi}}(O_t = o|O_{t-1}=o',S_t = s,B_t = b)\pi_{lo}^{\bm{\theta}_{lo}}(A_t=a|S_t=s,O_t = o).
\end{align*}
Exploiting the relation
\begin{align*}
    \mathbb{P}^{\bm{\theta}^{\text{old}}}_{o_0,s_1}(O_{t-1}=o',B_t=b,O_t=o,S_t=s,A_t=a|(s_t,a_t)_{1:T}) = \mathbb{P}^{\bm{\theta}^{\text{old}}}_{o_0,s_1}(O_{t-1}=o',B_t=b,O_t=o|(s_t,a_t)_{1:T})\delta(s-s_t)\delta(a-a_t)
\end{align*}
and rearranging the sums yields
\begin{align*}
    &Q_{o_0,s_1}^{T}(\bm{\theta} | \bm{\theta}^{\text{old}}) =\sum_{o'}\sum_b\sum_o\sum_s\sum_a\frac{1}{T}\sum_{t=1}^T\mathbb{P}^{\bm{\theta}^{\text{old}}}_{o_0,s_1}(O_{t-1}=o',B_t=b,O_t=o,S_t=s,A_t=a|(s_t,a_t)_{1:T}) \\
     &\times \log\pi_{b}^{\bm{\theta}_b}(B_t = b|S_t = s, O_{t-1}=o') \tilde \pi_{hi}^{\bm{\theta}_{hi}}(O_t = o|O_{t-1}=o',S_t = s,B_t = b)\pi_{lo}^{\bm{\theta}_{lo}}(A_t=a|S_t=s,O_t = o).
\end{align*}
Finally, we rewrite the probability in the previous equation as expectation of the indicator function
\begin{align*}
    &Q_{o_0,s_1}^{T}(\bm{\theta} | \bm{\theta}^{\text{old}}) =\sum_{o'}\sum_b\sum_o\sum_s\sum_a\frac{1}{T}\mathbb{E}_{o_0,s_1}^{\bm{\theta}^{\text{old}}}\bigg[ \sum_{t=1}^T\One[O_{t-1} = o', B_t = b, O_t = o, S_t=s, A_t=a] \bigg| (s_t,a_t)_{1:T}\bigg]\\
     &\times \log\pi_{b}^{\bm{\theta}_b}(B_t = b|S_t = s, O_{t-1}=o')\tilde \pi_{hi}^{\bm{\theta}_{hi}}(O_t = o|O_{t-1}=o',S_t = s,B_t = b)\pi_{lo}^{\bm{\theta}_{lo}}(A_t=a|S_t=s,O_t = o).
\end{align*}
Considering $\tilde \pi_{hi}^{\bm{\theta}_{hi}}$ in \eqref{eq:tilde_pi} and $\phi_{T}^{\bm{\theta}}$ in \eqref{eq:phi}, we obtain
\begin{equation*}
    \begin{split}
    Q_{o_0,s_1}^{T}(\bm{\theta} | \bm{\theta}^{\text{old}}) =& \sum_{o'}\sum_{o}\sum_s\sum_a \Big\{\sum_{b}\phi_{T}^{\bm{\theta}^{\text{old}}}(o', b, o, s, a)\log\pi_{b}^{\bm{\theta}_b}(b|s, o') \\
    &+\phi_{T}^{\bm{\theta}^{\text{old}}}(o', b = 1, o, s, a)\log \pi_{hi}^{\bm{\theta}_{hi}}(o|s) + \sum_{b}\phi_{T}^{\bm{\theta}^{\text{old}}}(o', b, o, s, a)\log \pi_{lo}^{\bm{\theta}_{lo}}(a|s,o)\Big\},
    \end{split}
\end{equation*}
which is \eqref{eq:Q_online_BW_Final}.

\subsection{Proof of Proposition~\ref{prop:OnlineBW_decomposition}}
\label{app:OnlineBW}
We start from the definition of $\rho_{T}^{\bm{\theta}}$ in \eqref{eq:OnlineBW_rho}:
\begin{align}
    \rho_{T}^{\bm{\theta}}(o',b,o,s,a,o'') =
        \frac{1}{T}\mathbb{E}^{\bm{\theta}}_{o_0,s_1}\bigg[ \sum_{t=1}^T\One[O_{t-1} = o', B_t = b, O_t = o, S_t=s, A_t=a] \bigg| O_T=o'', (s_t,a_t)_{1 : T}\bigg].
        \label{eq:app_rho}
\end{align}
Eq.~\eqref{eq:app_rho} is equivalent to
\begin{align}
\begin{split}
    \rho_{T}^{\bm{\theta}}(o',b,o,s,a,o'')
    = \frac{1}{T}\sum_{t=1}^T \mathbb{P}^{\bm{\theta}}_{o_0,s_1}(O_{t-1}=o', B_t=b, O_t=o, S_t=s, A_t=a|O_T=o'', (s_t,a_t)_{1:T}).
    \label{eq:app_prop1_rho_def}
\end{split}
\end{align}
Note that,
\begin{align}
    \begin{split}
    &\mathbb{P}^{\bm{\theta}}_{o_0,s_1}(O_{t-1}=o', B_t=b, O_t=o, S_t=s,A_t=a |O_T=o'', (s_t,a_t)_{1:T}) \\
    &= \frac{\mathbb{P}^{\bm{\theta}}_{o_0,s_1}(O_{t-1}=o', B_t=b, O_t=o, S_t=s,A_t=a,O_T=o'',  (s_t,a_t)_{1:T})}{\mathbb{P}^{\bm{\theta}}_{o_0,s_1}(O_T=o'', (s_t,a_t)_{1:T})}.
    \label{eq:app_prop1_P_joint}
    \end{split}
\end{align}
We plug \eqref{eq:app_prop1_P_joint} in \eqref{eq:app_prop1_rho_def} and break the sum in $t=T$ and $t<T$. Eq.~\eqref{eq:app_prop1_rho_def} becomes
\begin{align*}
    \rho_{T}^{\bm{\theta}}&(o',b,o,s,a,o'') =
    \frac{1}{T}\frac{ \mathbb{P}^{\bm{\theta}}_{o_0,s_1}(O_{T-1}=o', B_T=b,O_T=o,S_T=s,A_T=a, (s_t,a_t)_{1:T})}{\mathbb{P}^{\bm{\theta}}_{o_0,s_1}(O_T=o, (s_t,a_t)_{1:T})} \numberthis \label{eq:app_prop1_rho_first_part}\\
    &+ \bigg(1-\frac{1}{T}\bigg)\frac{1}{T-1}\sum_{t=1}^{T-1}\frac{\mathbb{P}^{\bm{\theta}}_{o_0,s_1}(O_{t-1}=o', B_t=b, O_t=o,S_t=s,A_t=a, O_T=o'', (s_t,a_t)_{1:T})}{\mathbb{P}^{\bm{\theta}}_{o_0,s_1}(O_T=o'', (s_t,a_t)_{1:T})}.\numberthis
    \label{eq:app_prop1_rho_second_part}
\end{align*}
Consider now the first term of the sum in \eqref{eq:app_prop1_rho_first_part}, it can be expanded as
\begin{align*}
    &\frac{1}{T}\frac{ \mathbb{P}^{\bm{\theta}}_{o_0,s_1}(O_{T-1}=o', B_T=b,O_T=o,S_T=s,A_T=a, (s_t,a_t)_{1:T})}{\mathbb{P}^{\bm{\theta}}_{o_0,s_1}(O_T=o, (s_t,a_t)_{1:T})}, \\
    &= \frac{1}{T}\frac{\delta(s-s_T)\delta(a-a_T)\pi_{lo}^{\bm{\theta}_{lo}}(a_T| s_T, o)\tilde \pi_{hi}^{\bm{\theta}_{hi}}(o|b,s_T,o')\pi_b^{\bm{\theta}_{b}}(b|s_T,o')\mathbb{P}^{\bm{\theta}}_{o_0,s_1}(O_{T-1}=o'|(s_t,a_t)_{1: T-1})}{\pi_{lo}^{\bm{\theta}_{lo}}(a_T | s_T, o)\sum_{o'}\sum_{b}\tilde \pi_{hi}^{\bm{\theta}_{hi}}(o|b,s_T,o')\pi_b^{\bm{\theta}_b}(b|s_T,o')\mathbb{P}^{\bm{\theta}}_{o_0,s_1}(O_{T-1}=o'|(s_t,a_t)_{1:T-1})}, \\
    &= \frac{1}{T}\frac{\delta(s-s_T)\delta(a-a_T)\tilde \pi_{hi}^{\bm{\theta}_{hi}}(o|b,s_T,o')\pi_b^{\bm{\theta}_{b}}(b|s_T,o')\mathbb{P}^{\bm{\theta}}_{o_0,s_1}(O_{T-1}=o'|(s_t,a_t)_{1: T-1})}{\sum_{o'}\sum_{b}\tilde \pi_{hi}^{\bm{\theta}_{hi}}(o|b,s_T,o')\pi_b^{\bm{\theta}_b}(b|s_T,o')\mathbb{P}^{\bm{\theta}}_{o_0,s_1}(O_{T-1}=o'|(s_t,a_t)_{1:T-1})}.\numberthis
    \label{eq:app_prop1_rho_first_part_final}
\end{align*}
For \eqref{eq:app_prop1_rho_second_part} instead, we first apply the total probability law with respect to $O_{T-1}=o'''$ and $B_T=b'$ and then proceed as in the previous term
\begin{align*}
    &\bigg(1-\frac{1}{T}\bigg)\frac{1}{T-1}\sum_{t=1}^{T-1}\sum_{o'''}\sum_{b'}\frac{ \mathbb{P}^{\bm{\theta}}_{o_0,s_1}(O_{t-1}=o', B_t=b, O_t=o, S_t=s, A_t=a, O_{T-1} = o''',B_T = b', O_T = o'', (s_t,a_t)_{1:T})}{\mathbb{P}^{\bm{\theta}}_{o_0,s_1}(O_T = o'', (s_t,a_t)_{1:T})} \\
    =&\bigg(1-\frac{1}{T}\bigg)\frac{1}{T-1}\sum_{t=1}^{T-1}\sum_{o'''}\sum_{b'}\mathbb{P}^{\bm{\theta}}_{o_0,s_1}(O_{t-1}=o', B_t=b, O_t=o,S_t=s, A_t=a, O_{T-1} = o'''|(s_t,a_t)_{1:T-1})
    \\
    &\times \frac{\tilde \pi_{hi}^{\bm{\theta}_{hi}}(o''|b',s_T,o''')\pi_b^{\bm{\theta}_b}(b'|s_T,o''')}{\sum_{o'''}\sum_{b'}\tilde \pi_{hi}^{\bm{\theta}_{hi}}(o''|b',s_T,o''')\pi_b^{\bm{\theta}_b}(b'|s_T,o''')\mathbb{P}^{\bm{\theta}}_{o_0,s_1}(O_{T-1} = o'''|(s_t,a_t)_{1:T-1})}
    \\
        =&\bigg(1-\frac{1}{T}\bigg)\sum_{o'''}\sum_{b'}\frac{1}{T-1}\sum_{t=1}^{T-1}\mathbb{P}^{\bm{\theta}}_{o_0,s_1}(O_{t-1}=o', B_t=b, O_t=o, S_t=s, A_t=a| O_{T-1} = o''', (s_t,a_t)_{1:T-1}) \\
        &\times \frac{\tilde \pi_{hi}^{\bm{\theta}_{hi}}(o''|b',s_T,o''')\pi_b^{\bm{\theta}_b}(b'|s_T,o''')\mathbb{P}^{\bm{\theta}}_{o_0,s_1}(O_{T-1} = o'''|(s_t,a_t)_{1: T-1})}{\sum_{o'''}\sum_{b'}\tilde \pi_{hi}^{\bm{\theta}_{hi}}(o''|b',s_T,o''')\pi_b^{\bm{\theta}_b}(b'|s_T,o''')\mathbb{P}^{\bm{\theta}}_{o_0,s_1}(O_{T-1} = o'''|(s_t,a_t)_{1:T-1})}
 \\
        =&\bigg(1-\frac{1}{T}\bigg)\sum_{o'''}\sum_{b'}\rho_{T-1}^{\bm{\theta}}(o', b, o, s, a, o''') \\
        &\times \frac{\tilde \pi_{hi}^{\bm{\theta}_{hi}}(o''|b',s_T,o''')\pi_b^{\bm{\theta}_b}(b'|s_T,o''')\mathbb{P}^{\bm{\theta}}_{o_0,s_1}(O_{T-1} = o'''|(s_t,a_t)_{1: T-1})}{\sum_{o'''}\sum_{b'}\tilde \pi_{hi}^{\bm{\theta}_{hi}}(o''|b',s_T,o''')\pi_b^{\bm{\theta}_b}(b'|s_T,o''')\mathbb{P}^{\bm{\theta}}_{o_0,s_1}(O_{T-1} = o'''|(s_t,a_t)_{1:T-1})}. \numberthis
        \label{eq:app_prop1_rho_second_part_final}
\end{align*}
By summation of \eqref{eq:app_prop1_rho_first_part_final} and \eqref{eq:app_prop1_rho_second_part_final} we obtain the final recursion for $\rho_T^{\bm{\theta}}$ in Proposition~\ref{prop:OnlineBW_decomposition}
\begin{align*}
    \begin{split}
    &\rho_{T}^{\bm{\theta}}(o', b, o, s, a, o'')
    = \sum_{o'''}\sum_{b'}\bigg\{\frac{1}{T}\kappa(o'-o''', b-b', o-o'', s-s_T, a-a_T) \\ &+  \bigg(1-\frac{1}{T}\bigg)\rho_{T-1}^{\bm{\theta}}(o', b, o, s, a, o''')\bigg\} \frac{\tilde\pi_{hi}^{\bm{\theta}_{hi}}(o''|b',s_T,o''')\pi_b^{\bm{\theta}_{b}}(b'|s_T,o''')\mathbb{P}^{\bm{\theta}}_{o_0,s_1}(O_{T-1} = o'''|(s_t,a_t)_{1: T-1})}{\sum_{o'''}\sum_{b'}\tilde \pi_{hi}^{\bm{\theta}_{hi}}(o''|b',s_T,o''')\pi_b^{\bm{\theta}_{b}}(b'|s_T,o''')\mathbb{P}^{\bm{\theta}}_{o_0,s_1}(O_{T-1} = o'''|(s_t,a_t)_{1: T-1})},
\end{split}
\end{align*}
where $\kappa(o'-o''', b-b', o-o'', s-s_T, a-a_T)=\delta(o'-o''')\delta(b-b')\delta(o-o'')\delta(s-s_T)\delta(a-a_T)$.
In a similar way, we show the recursion for $\chi_{T}^{\bm{\theta}}(o) = \mathbb{P}^{\bm{\theta}}_{o_0,s_1}\big(O_T = o\big|(s_t,a_t)_{1 : T}\big)$
\begin{align*}
    &\mathbb{P}^{\bm{\theta}}_{o_0,s_1}\big(O_T = o\big|(s_t,a_t)_{1:T}\big) =\sum_{o'}\sum_{b} \frac{\mathbb{P}^{\bm{\theta}}_{o_0,s_1}(O_{T-1} = o', B_T = b, O_T = o,(s_t,a_t)_{1: T})}{\mathbb{P}^{\bm{\theta}}_{o_0,s_1}((s_t,a_t)_{1: T})}\\
    &= \sum_{o'}\sum_{b} \frac{\pi_{lo}^{\bm{\theta}_{lo}}(a_T | s_T, o)\tilde \pi_{hi}^{\bm{\theta}_{hi}}(o|b,s_T,o')\pi_b^{\bm{\theta}_b}(b|s_T,o')\mathbb{P}^{\bm{\theta}}_{o_0,s_1}(O_{T-1} = o'|(s_t,a_t)_{1: T-1})}{\sum_{o'}\sum_{b}\sum_{o}\pi_{lo}^{\bm{\theta}_{lo}}(a_T | s_T, o)\tilde \pi_{hi}^{\bm{\theta}_{hi}}(o|b,s_T,o')\pi_b^{\bm{\theta}_b}(b|s_T,o')\mathbb{P}^{\bm{\theta}}_{o_0,s_1}(O_{T-1} = o'|(s_t,a_t)_{1 : T-1})}.
\end{align*}
We end up with the recursion in Proposition~\ref{prop:OnlineBW_decomposition}
\begin{align*}
    \begin{split}
        &\chi_{T}^{\bm{\theta}}(o)
        = \sum_{o'}\sum_{b}\frac{\pi_{lo}^{\bm{\theta}_{lo}}(a_T | s_T, o)\tilde \pi_{hi}^{\bm{\theta}_{hi}}(o|b,s_T,o')\pi_b^{\bm{\theta}_{b}}(b|s_T,o')\chi_{T-1}^{\bm{\theta}}(o')}{\sum_{o'}\sum_{b}\sum_{o}\pi_{lo}^{\bm{\theta}_{lo}}(a_T | s_T, o)\tilde \pi_{hi}^{\bm{\theta}_{hi}}(o|b,s_T,o')\pi_b^{\bm{\theta}_{b}}(b|s_T,o')\chi_{T-1}^{\bm{\theta}}(o')}
    \end{split}
\end{align*}

\subsection{Regularization Penalties}
\label{app:regularizers}

The penalties in \eqref{eq:pi_hi_reg} and \eqref{eq:regularizer_KL_div} are differently implemented in the batch and in the online algorithm,

\emph{Batch BW (Algorithm~\ref{alg:batchBW}):} 
\begin{align*}
    \begin{split}
    L_b &= \sum_{o_t} \big|\big| \mathbb{E}_{s_t}[\pi_{hi}^{\bm{\theta}_{hi}}(o_t|s_t)] - \tau \big|\big|_{2} \approx \sum_{o_t}\big|\big|\frac{1}{T}\sum_{j=1}^T\pi_{hi}^{\bm{\theta}_{hi}}(o_t|s_j) - \tau \big|\big|_2, 
    \end{split} \\
    \begin{split}
    L_v &= \sum_{o_t} \text{var}_{s_t}[\pi_{hi}^{\bm{\theta}_{hi}}(o_t|s_t)] \approx \sum_{o_t} \frac{1}{T}\sum_{j=1}^T\Big(\pi_{hi}^{\bm{\theta}_{hi}}(o_t|s_j) - \big( \frac{1}{T}\sum_{i=1}^T\pi_{hi}^{\bm{\theta}_{hi}}(o_t|s_i)\big)\Big)^2, 
    \end{split} \\
    \begin{split}
      L_{D_{KL}} &= \sum_{o}\sum_{o',o\neq o'} D_{KL}(\pi_{lo}(a_t|s_t,o)||\pi_{lo}(a_t|s_t,o'))
      \approx \sum_{o}\sum_{o',o\neq o'}\bigg\{\frac{1}{T}\sum_{t=1}^T\pi_{lo}(a_t|s_t,o)\log\frac{\pi_{lo}(a_t|s_t,o)}{\pi_{lo}(a_t|s_t,o')}\bigg\}. 
    \end{split}
\end{align*}
\emph{Online BW (Algorithm~\ref{alg:onlineBW_ass2}):}
\begin{align*}
    \begin{split}
    L_b &= \sum_{o} \big|\big| \mathbb{E}_{s}[\pi_{hi}^{\bm{\theta}_{hi}}(o|s)] - \tau \big|\big|_{2} \approx 
    \sum_{o}\big|\big|\frac{1}{|\mathcal{\tilde S}|}\sum_{s}\pi_{hi}^{\bm{\theta}_{hi}}(o|s) - \tau \big|\big|_2, 
    \end{split} \\
    \begin{split}
    L_v &= \sum_{o} \text{var}_{s}[\pi_{hi}^{\bm{\theta}_{hi}}(o|s)] \approx \sum_{o} \frac{1}{|\mathcal{\tilde S}|}\sum_{s_j}\Big(\pi_{hi}^{\bm{\theta}_{hi}}(o|s_j) - \big(\frac{1}{|\mathcal{\tilde S}|}\sum_{s_i}\pi_{hi}^{\bm{\theta}_{hi}}(o_t|s_i)\big)\Big)^2, 
    \end{split} \\
    \begin{split}
      L_{D_{KL}} &= \sum_{o}\sum_{o',o\neq o'} D_{KL}(\pi_{lo}(a_T|s_T,o)||\pi_{lo}(a_T|s_T,o'))
      \approx \sum_{o}\sum_{o',o\neq o'}\bigg\{\pi_{lo}(a_T|s_T,o)\log\frac{\pi_{lo}(a_T|s_T,o)}{\pi_{lo}(a_T|s_T,o')}\bigg\}. 
    \end{split}
\end{align*}

\subsection{Experiments Details}
\label{app:experiments}
Table~\ref{tab:app_hyper} shows the hyperparameters we use for all the environments, trials and seeds on the experiments in Section~\ref{sec:experiments}.
\begin{table}[h!]
\centering
\begin{tabular}{ |c|l|l|l|  }
 \hline
 \multicolumn{4}{|c|}{Hyperparameters} \\
 \hline
  & & Batch & Online \\
 \hline
 {Fig.~\ref{fig:Experiments}} & $\lambda_b$ & 1 & 1 \\
 & $\lambda_v$ & 0.1 & 0.1 \\
 & $\lambda_{D_{KL}}$ & 0.01 & 0.01 \\
 & $N$ & 20 & N.A.  \\
 & Optimizer & Adamax & Adamax \\
 & Optimization Algorithm & Mini-Batch Gradient Descent & Gradient Descent \\
 & Mini-Batch size & 32 & N.A.\\
 & Full Gradient steps per iterations & 50 & 30 \\
 & Learning rate & $10^{-2}$ & $10^{-2}$ \\
 \hline
\end{tabular}
\caption{Table of hyperparameters used for Section~\ref{sec:experiments}}
\label{tab:app_hyper}
\end{table}

\end{appendix}

\end{document}